\documentclass[letterpaper]{article} %DO NOT CHANGE THIS
\usepackage{aaai18}  %Required
\usepackage{times}  %Required
\usepackage{helvet}  %Required
\usepackage{courier}  %Required
\usepackage{url}  %Required
\usepackage{graphicx}  %Required
\usepackage{subcaption}
\usepackage{verbatim}  %Required
\usepackage{color}
\usepackage{enumitem}
\nocopyright{}
\begin{document}
    % The file aaai.sty is the style file for AAAI Press 
    % proceedings, working notes, and technical reports.
    %
    
    %% \todo{} command.
    %
    % Outputs red TODOs in the document. Requires \usepackage{color}.
    %
    % Usage: \todo{Document the TODO command.}
    %
    % Comment out second line to disable.
    \newcommand{\todo}[1]{}
    \renewcommand{\todo}[1]{{\color{red} TODO: {#1}}}
    
    \title{Interaction and Autonomy in RoboCup@Home and Building-Wide Intelligence}
    
    \author{
        Justin Hart\textsuperscript{1} \and Harel Yedidsion\textsuperscript{1} \and Yuqian Jiang\textsuperscript{1} \and Nick Walker\textsuperscript{2} \and \\
        \textbf{\Large{Rishi Shah\textsuperscript{1} \and Jesse Thomason\textsuperscript{2} \and Aishwarya Padmakumar\textsuperscript{1} \and Rolando Fernandez\textsuperscript{1} \and}} \\
        \textbf{\Large{ Jivko Sinapov\textsuperscript{3} \and Raymond Mooney\textsuperscript{1} \and Peter Stone\textsuperscript{1}}}\\
        \textsuperscript{1}Department of Computer Science, University of Texas at Austin, Austin, Texas, USA\\
        \textsuperscript{2}Paul G. Allen School of Computer Science \& Engineering, University of Washington, Seattle, Washington, USA\\
        \textsuperscript{3}Department of Computer Science, Tufts University, Medford, Massachusetts, USA\\
        \{hart, harel, rishi, aish, mooney, pstone\}@cs.utexas.edu,
        \{jiangyuqian, rfernandez\}@utexas.edu, \\
        \{nswalker, jdtho\}@cs.washington.edu, jsinapov@cs.tufts.edu
    }

    \maketitle

    \begin{abstract}
        Efforts are underway at UT Austin to build autonomous robot systems that address the challenges of long-term deployments in office environments and of the more prescribed domestic service tasks of the RoboCup@Home competition. We discuss the contrasts and synergies of these efforts, highlighting how our work to build a RoboCup@Home Domestic Standard Platform League entry led us to identify an integrated software architecture that could support both projects. Further, naturalistic deployments of our office robot platform as part of the Building-Wide Intelligence project have led us to identify and research new problems in a traditional laboratory setting.
    \end{abstract}
    
    \section{Introduction}
    
    Pursuing research on multiple, related fronts can lead you to challenges that would be hidden by viewing problems from only one perspective. The experience of working simultaneously on multiple service robot projects -- RoboCup@Home, the Building-Wide Intelligence (BWI) Project, and our laboratory experiments -- has become a guiding force in our group's efforts over the past year. As a principle, we have chosen approaches that we envision as solving not only short-term goals on these projects, but ones that we identify as immediate obstacles to comprehensive general purpose service robots that cross these domains. Service robots are envisioned to soon aid non-expert users in a variety of tasks. The realization of mature service robot technologies will involve developments in both Artificial Intelligence (AI) and Human-Robot Interaction (HRI). By working toward a comprehensive system that is capable of winning RoboCup@Home, is used by the occupants of the Computer Science Department at UT Austin, and is used in our laboratory experiments, we have explored a number of important problems in AI and HRI.
    
    In this paper, we discuss our efforts toward such a comprehensive system. RoboCup@Home evaluates domestic service robots on a variety of tasks in a simulated apartment. The BWI project aims to deploy office assistant robots that accomplish useful tasks for the occupants of UT Austin's computer science building. Our laboratory experiments have focused on solutions to AI and HRI problems contributing to the development of such systems. As a result, by 2017 BWI had already developed a mature suite of software for processing natural language instructions, for navigation, and for semantic representations allowing the robot to formulate plans involving locations and objects. Leveraging this infrastructure helped us to quickly develop a system for competing in RoboCup@Home. However, a desire to develop a comprehensive system capable of both solving all of the stages of RoboCup@Home and fitting the needs of BWI required us to reconsider these capabilities, leading to the development of an updated Knowledge Representation and Reasoning (KR\&R) system and a top-level architecture that allows us to perform planning, perception, and action, while leveraging a core system that can be tailored to each specific task in the competition.
    
    A key realization that we had while pursuing RoboCup@Home is that an architecture like ours needs a rich semantic representation of the environment that the robot operates in. Bringing lessons learned back into the BWI project -- taking this out of a simulated apartment and back into the real world -- poses a significant challenge in terms of both scope and scale. This led us to work on Pose Registration for Integrated Semantic Mapping (PRISM), a system for producing semantic map annotations based on the robot's percepts.
    
    Fielding a live system in the real world has also led us to important conclusions based on how people interact with the robot, and there is an interplay between the challenges of HRI in the RoboCup@Home environment and those in the BWI environment. Navigating the, at times crowded, corridors of the building has led us to studies in HRI for navigational tasks, but also led to discoveries in how people develop an understanding of signaling mechanisms. Challenges like identifying restaurant patrons or who, from a group of people, is talking to the robot, will help to drive further future developments on our system.
    
    \begin{figure}
        \centering
        \includegraphics[width=0.45\textwidth]{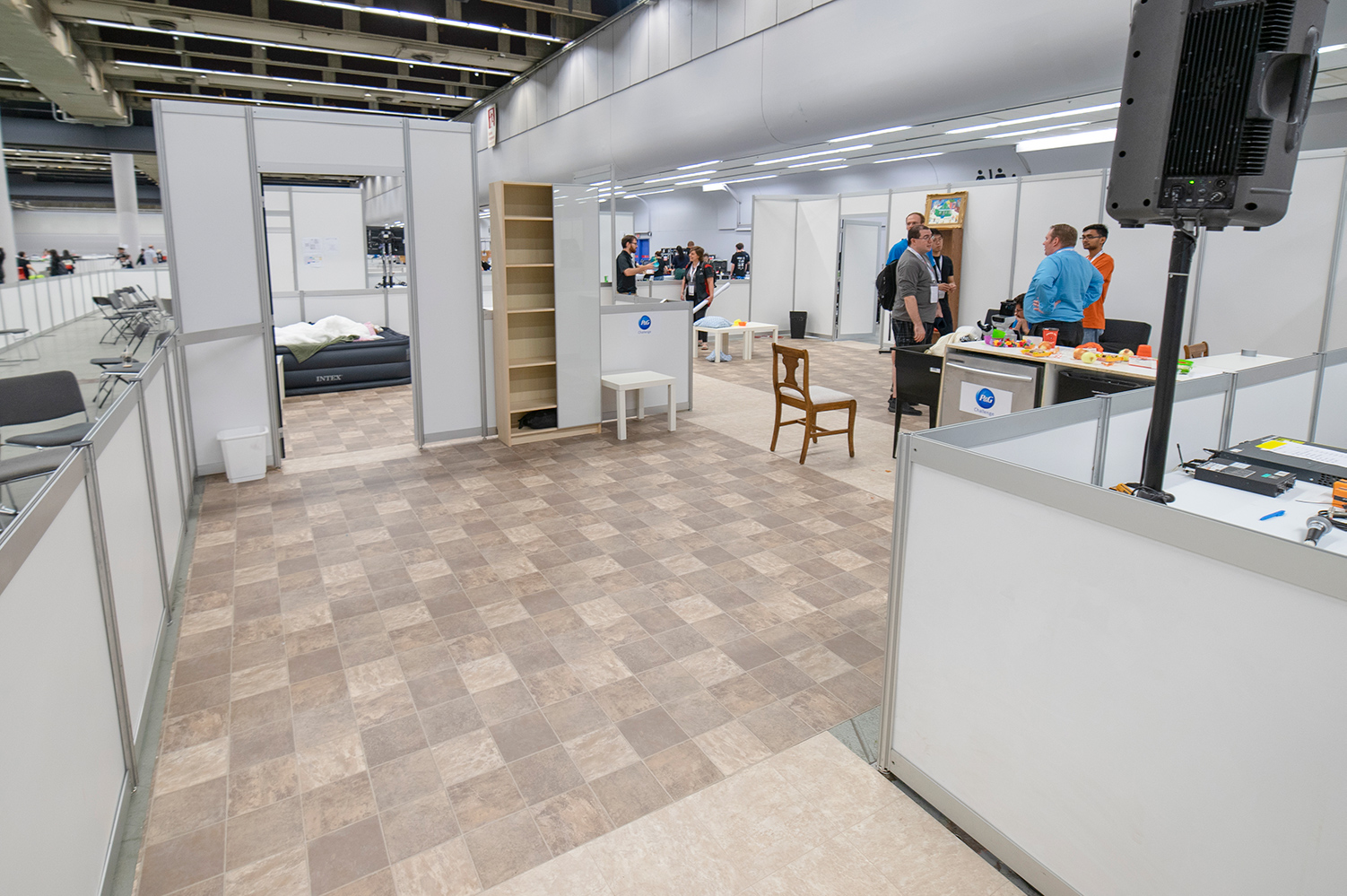}
        \caption{RoboCup@Home Arena for the Domestic Standard Platform League in the 2018 competition in Montreal.}
        \label{fig:robocup_arena}
    \end{figure}
    \section{From BWI to RoboCup@Home}
    
\begin{figure}
        \centering
        \begin{subfigure}{0.45\columnwidth}
            \includegraphics[width=\columnwidth]{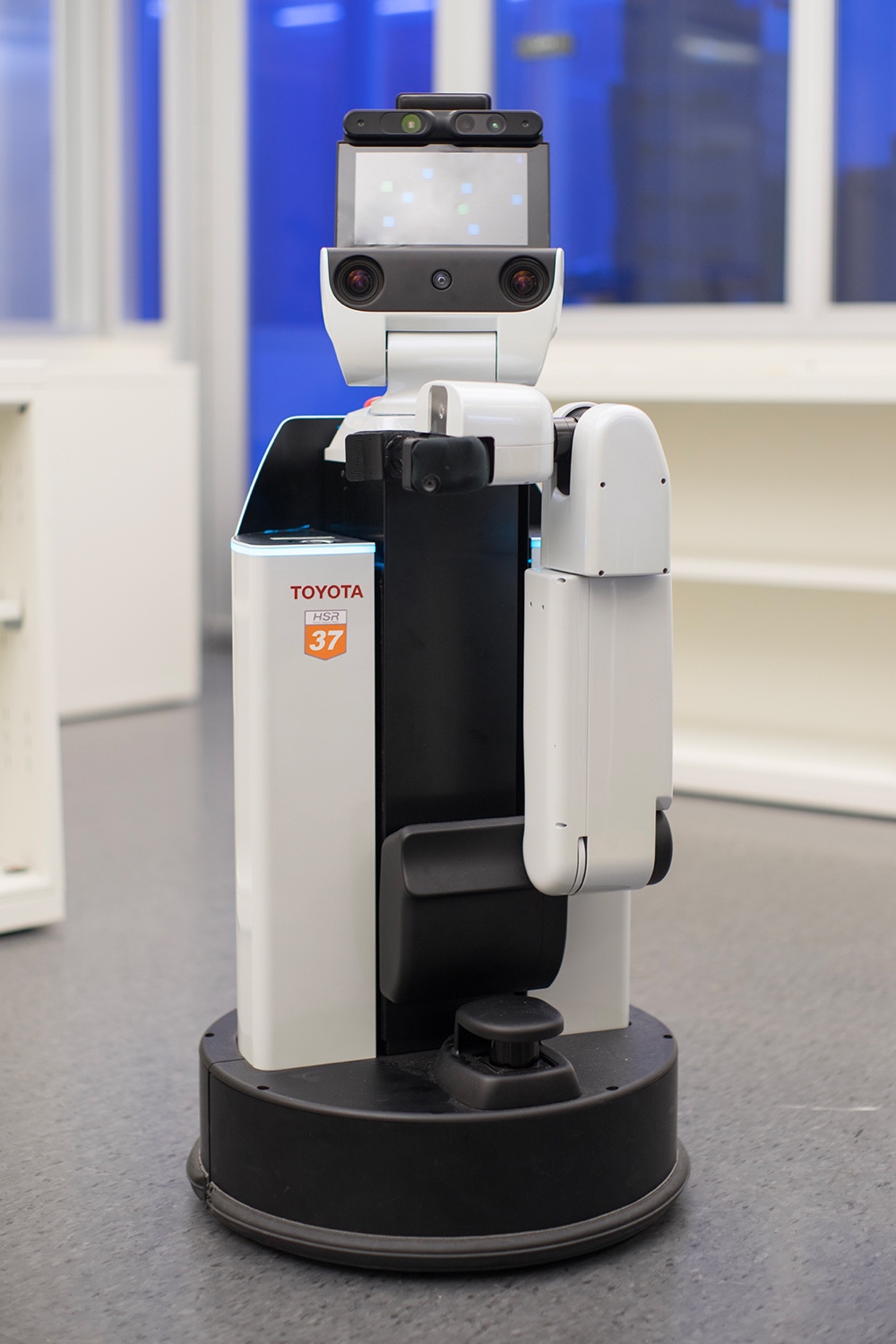}	
            \caption{Toyota HSR.}	
            \label{fig:hsr}
        \end{subfigure}
        \quad
        \begin{subfigure}{0.45\columnwidth}
            \includegraphics[width=\columnwidth]{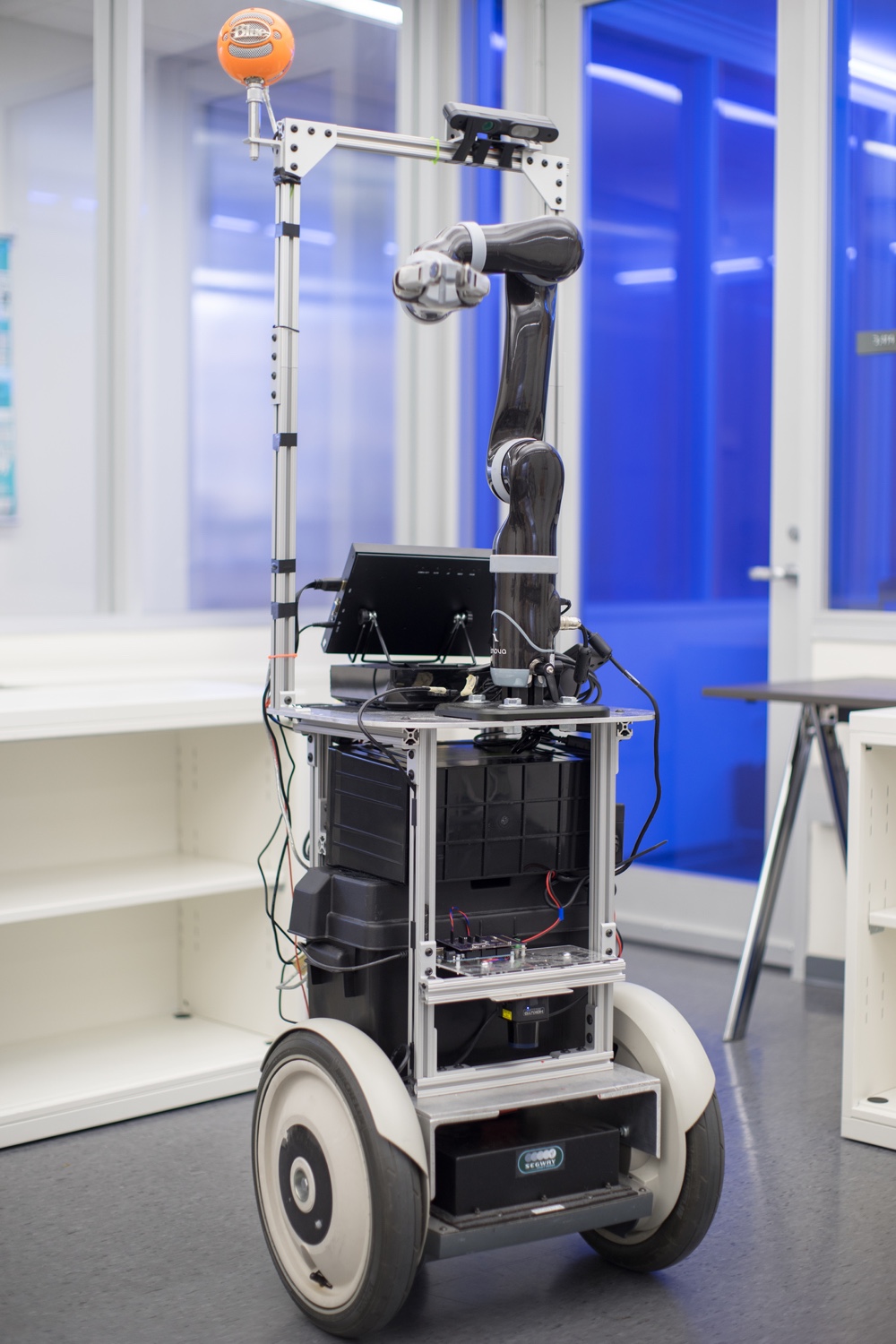}
            \caption{BWIBot.}	
            \label{fig:bwi_bot}
        \end{subfigure}	
        \caption{The BWIBot and HSR are similar robots from a functional perspective. Both have a single arm, a Asus Xtion Pro RGB-D camera, and highly-maneuverable bases.}	
        \label{fig:robots}
    \end{figure}

    RoboCup@Home is set in a simulated apartment space (called the arena), as can be seen in Figure \ref{fig:robocup_arena}. There are three leagues, each based around a different robot platform. Our team, UT Austin Villa@Home, competes in the Domestic Standard Platform League (DSPL), which uses the Toyota Human Support Robot (HSR), as seen in Figure \ref{fig:hsr}.
    
    Each RoboCup@Home competition consists of tests in two stages. The Stage I tasks are \textit{General Purpose Service Robot} (GPSR), in which commands are given in natural language (generated by a grammar), and can require the robot to leverage any skill found in other tasks; \textit{Speech and Person Recognition}, where the robot plays question-answering games and is scored partly for looking at the participant who asked the question;
    \textit{Storing Groceries}, in which the robot takes household items from a kitchen table and stores them in a cupboard near other similar items; and \textit{Help-Me-Carry}, where the robot follows a human operator to a car, carries groceries back, and leads a person in the arena back out to carry more groceries.
    
\begin{figure}
        \centering
        \includegraphics[width=0.45\textwidth]{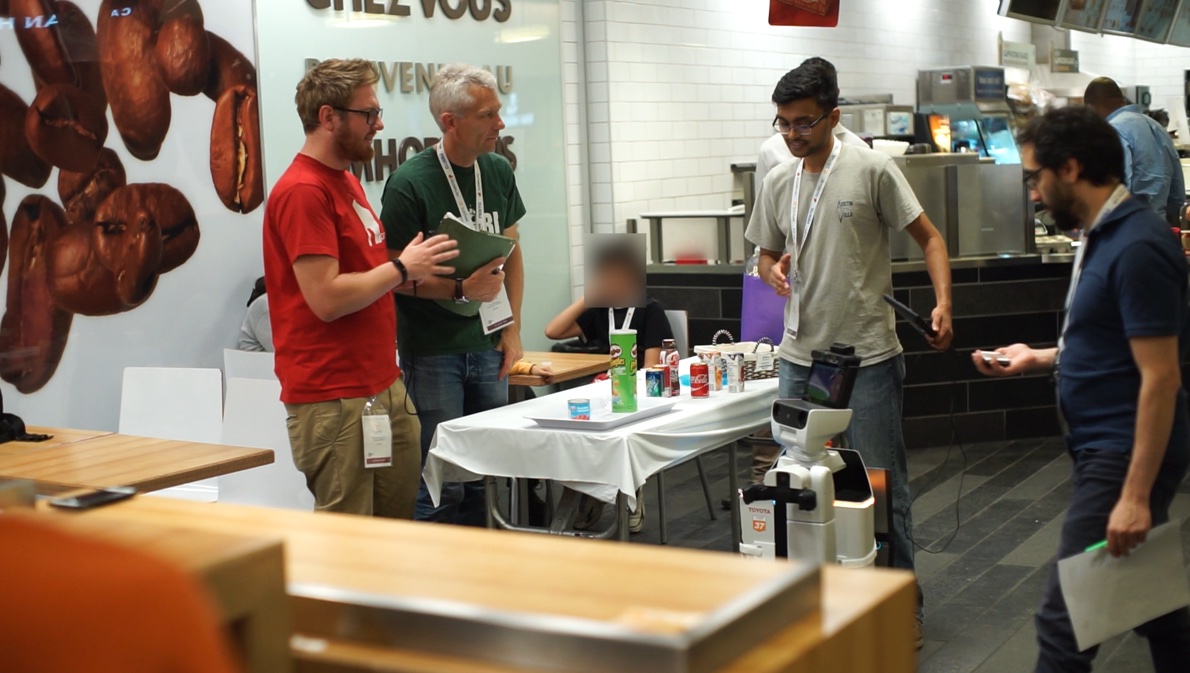}
        \caption{Restaurant Task from the 2018 competition.}
        \label{fig:restaurant}
    \end{figure}

    Stage II tasks include \textit{Enhanced Endurance General Purpose Service Robot} (EEGPSR), a longer version of GPSR with more complex commands which may contain incomplete or false information; \textit{Procter \& Gamble Dishwasher Challenge}, where the robot cleans a table and loads dishes into a dishwasher; and \textit{Restaurant} where patrons call to the robot, which takes orders and goes to a bar area to fulfill them. The task is hosted in a real restaurant. In 2018, restaurant was the coffee shop at the convention center, seen in Figure \ref{fig:restaurant}.

    \subsection{Leveraging BWI Infrastructure}
    
    RoboCup@Home is a competition which tests domestic service robots for their ability to perform a variety of tasks representative of what home users would soon like to see in a real product. Building-Wide Intelligence, on the other hand, is a real-world deployment of a robot in which we attempt to flexibly respond to commands given to the robot by both regular and new users interacting with it on a day-to-day basis in our Computer Science department.
    
    Our group became interested in participation in RoboCup@Home due to parallels between our ongoing research and the tasks posed to teams in this competition. BWI projects included work on knowledge reasoning and planning \cite{khandelwal2014planning}, extensive work in natural language processing and perceptual grounding \cite{IJCAI16-thomason,thomason2017opportunistic,thomason:mrhrc18}, and multi-robot human guidance \cite{Khandelwal:2017:MHG:3091125.3091314}.
    
    %\subsection{Natural Language Interaction and Planning}
    Significant research on the BWI project has been dedicated to the problem of robots interacting with humans using natural language. This has included work on the perceptual grounding of language \cite{thomason2017opportunistic}, dialog strategies \cite{padmakumar_eacl17}, and semantic parsing \cite{Thomasonthesis}. In recent work, the robot acts on verbal commands from a human interlocutor \cite{thomason:ijcai15,thomason:mrhrc18}. In particular, this vein of research has direct parallels in RoboCup@Home. In recent work \cite{thomason:mrhrc18}, a BWIBot is instructed to find an object and bring it to a person's office. This has parallels in  the General Purpose Service Robot (GPSR) and Enhanced Endurance General Purpose Service Robot (EEGPSR) RoboCup@Home tasks, in which a human operator gives tasking to the robot as verbal commands.
    
    Our existing research on BWI, and the similarity of the project to the stages of RoboCup@Home enabled us to quickly ramp up our team for competition. Though superficially different, the BWIBot and the Toyota HSR, used in the RoboCup@Home Domestic Standard Platform League (DSPL) are remarkably similar. They can be seen in Figure \ref{fig:robots}. Both have one arm, Asus Xtion Pro cameras, speech input capabilities, and highly maneuverable bases. In our first year participating in the RoboCup@Home, we mostly leveraged existing software from BWI, and took third place after only about two months of preparation.

    \subsection{A Complete Approach to RoboCup@Home}
    Our goal for RoboCup@Home is a to build a single, unified system which is capable of competing in every round of RoboCup@Home as well as running BWI. The complexity of accomplishing the multiple tasks represented in RoboCup@Home lends itself to an approach wherein teams write custom software for each task. Successfully constructing such a system would represent a real general purpose service robot which is capable of flexibly completing a variety of tasks. To approach this goal, we developed first enhanced our software architecture and knowledge representation capabilities.
    
    \subsubsection{Architecture}
    Our new robot architecture is composed of three layers: a reactive layer at the top, a deliberative control module, and skills. The top layer in this architecture is composed of Hierarchical Finite State Machines (FSMs) implemented in SMACH \cite{smach}. This gives us two fundamental capabilities: to specify a step-by-step top-level script of the robot's behavior, and to quickly and reactively control the robot - especially under error conditions. The middle layer, with a planner and plan executor, allows the robot to reason over its knowledge base in order to formulate its own plans as necessary. This can be bypassed by the top layer as merited. For instance, the robot may have a state machine which instructs it to wander hallways looking for people to interact with, but a planner which takes over when given a spoken command. At the bottom level are skills, low-level behaviors that the robot can engage, such as sensing or manipulation. They give our architecture the flexibility to leverage existing software such as interfaces written in ROS or machine learning code implemented as neural networks.
    
    \subsubsection{Knowledge Representation}
    
    As discussed above, existing BWI software is capable of grounding objects in interaction and service object retrieval commands. In the GPSR and EEGPSR tests of RoboCup@Home, the robot is required to manipulate unknown objects and interact with unseen people, and report errors if the objects or people cannot be found. For instance, the command "bring me an apple from the kitchen" implies that the operator thinks there is an apple in the kitchen, but the robot might not have sensed this specific apple, and there might not be an apple in the kitchen at all. This challenge motivates expanding our knowledge representation system to handle planning and reasoning in the open world, as well as representing hypothetical information from the operator. 
    
    Our solution leverages semantic networks in a knowledge base to represent object relations and attributes, including if they are hypothetical. The current knowledge, as well as hypotheses of ungrounded objects are injected into the reasoning system to generate plans. For example, to solve for the command "bring me an apple from the kitchen", our system assumes there is a hypothetical apple that can be found at any location in the kitchen, and generates a plan to one of them. If the robot cannot find an apple at the location, the system replans to the next possible location. In the case that there is no feasible plan because the robot has searched all locations, our system triggers diagnostics that reason about the hypotheses implied by the operator. Since the robot believes there is no apple at any location in the kitchen, the hypothesis that there is an apple in the kitchen becomes invalid.
    
    \section{From RoboCup@Home to BWI}
    This system is now being ported back to the BWIBot to allow us to run a unified architecture across both platforms, and to use what we have learned at RoboCup@Home in BWI. Our plan for BWI going into Fall 2018 is to create a top-level state machine which describes the basic behavior of the BWIBot, with a variety of basic interactions which can be instantiated from there. Individual, brief interactions can implement capabilities such as providing directions to a person, object delivery, or other tasks, as the situation merits. The top-level architecture provides a means to gracefully switch between these behaviors, while providing a framework for both long-term autonomy and the flexibility to deploy experimental systems.
    
    \subsection{Semantic Mapping}
    Another branch of our research which has been inspired by RoboCup@Home work is the autonomous semantic labeling of maps. An important skill in RoboCup@Home is locating objects and navigating with respect to human-recognizable landmarks such as rooms in the arena, furniture, and appliances. Part of how teams are able to perform complex verbal commands is due to the ability to manually add a subset of the objects the robot interacts with to their object recognition subsystems, and to spend ``setup days'' manually annotating poses and areas in the RoboCup@Home arena. Commands may be akin to, ``Bring the juice from the living room to Jan, who is in the bedroom.'' Eventually, we would like to be able to start the robot at the RoboCup@Home arena and have it autonomously create maps. Bringing these skills back to BWI also entails the challenge that creating such manual annotations does not scale to large buildings such as our Computer Science department, and that such maps could not be kept up-to-date in a building with hundreds of occupants. We would like to uncover the relevant semantic information on-the-fly during competition at RoboCup@Home and during operation in BWI.
    
    To approach this problem, we have been building a system called ``Pose Registration for Integrated Semantic Mapping'' or ``PRISM'' to help accomplish these tasks. The basic technology starts with a system that is able to run SLAM to create a navigational map of the environment. To this map, we add pose registrations of objects found in the environment and annotate the semantics of these objects into the robot's knowledge base.
    
    The basic system is built from three components: a classifier, which recognizes objects to be incorporated into the map; a pose estimator, which estimates their pose to be annotated on the map; and an extractor, which extracts semantic information to be annotated. 
    
    Our first efforts on this front have been to implement software which allows our robot to extract text from the signage in our building in order to label this information into its navigational map \cite{justin_iros_2018}. A custom classifier scans for objects which look like the office placards in our building. Since they are planar targets, their pose can be determined by computing a homography with respect to them. This homography can also be used to rectify the image, aiding in the extraction of text to annotate the placard's semantics into the robot's knowledge base.
    
    To expand this system, we are now both adding different types of map annotations that the robot can make, and working on enabling the robot to autonomously perform an exploration of its building in order to create these maps.
    
    \section{Fielding Real Robots Reveals Research Problems}
    The BWIBots are intended to be a live deployment of a real fleet of general purpose service robots for use in our Computer Science department. As such, we run the robots continually in our building. One problem that we have run into, however, is that uncertainty between the robot and the human concerning the direction in which they intend to navigate can create traffic issues in our hallways. Imagine the scenario in which you walk towards another person, intending to walk past, but take a right step, and they take a left step, resulting in you blocking each other's paths. Our robots were facing just this problem.
    
    \subsection{LED Turn Signals}
\begin{figure}
        \centering
        \includegraphics[width=0.45\textwidth]{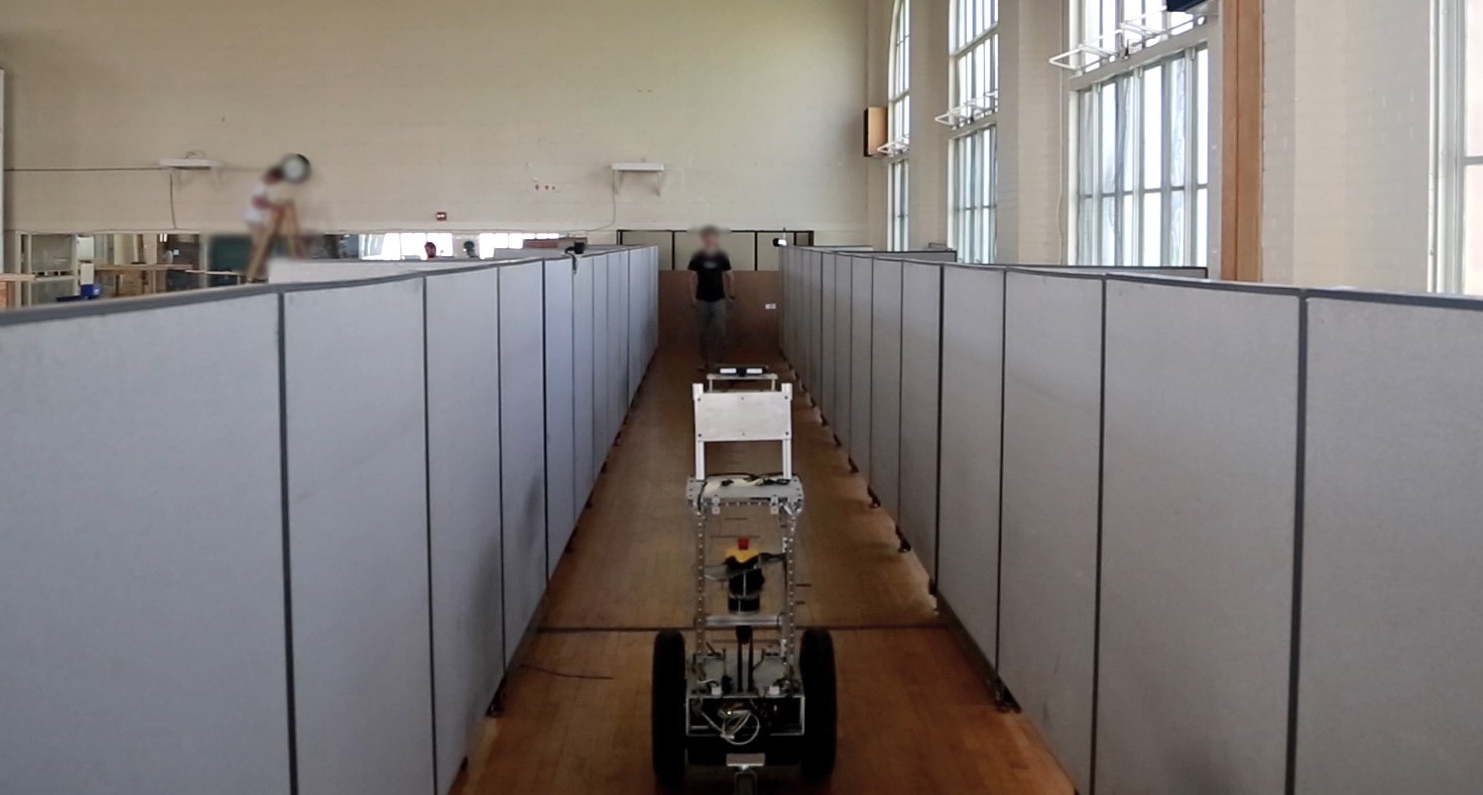}
        \caption {Constructed hallway environment with robot and participant in the early stage of hallway traversal.}
        \label{fig:hallway}
    \end{figure}

    To study this problem, we constructed a $17.5 \times 1.85$m hallway from cubicle furniture in which to test approaches to disambiguating the robot's intended path, Figure \ref{fig:hallway}, allowing for a person navigating in the direction toward the robot to easily pass. We tested the idea of using LED turn signals mounted to the robot to disambiguate its navigational intention.
    
    To assure that study participants reacted to the robot's turn signal, rather than its motion, we empirically determined distances and speeds which would assure that the human and robot would come into conflict with each other if the person did not react to the robot's signal. The robot would always turn left, blocking the person's path if they turned to the right. This runs counter to people's intuitions, which is to pass each other in the hallway by moving to the right. The robot begins its turn at $2.75$m, which is too close for the person to change paths to avoid the robot. If the robot comes within $1$m of the person, it comes to a complete stop. Stops, or behaviors where the person turns into the robot's path, but manages to avoid its motion, are marked as conflicts. What we found from this study, however, is that study participants were unable to understand the robot's turn signals, and so they were unhelpful in helping to resolve the navigational conflict.
    
    \subsection{Passive Demonstrations}
    We wanted to instruct users as to the intention of the turn signal without an explicit period of user training. To this end, we introduce the concept of a \textit{passive demonstration}. \cite{rolando_roman_2018} In a passive demonstration, the robot uses the signal in context, demonstrating its intent, before it is needed to resolve a real conflict.
    
    In our experiment, the robot makes a turn, using its turn signal, at the very start of its motion at the opposite end of the test hallway. The participant can then see the signal in use before coming close enough to the robot for its usage to matter in resolving their navigational conflict.
    
    We found was that passive demonstrations dramatically improved performance. With no demonstration, while using the LEDs, the robot conflicted with participants $90\%$ of the time. With the demonstration, however, this went down to $20\%$. We are currently investigating approaches to changing the robots motion, to interpreting human motion, and to incorporating signals with passive demonstrations into our robot, all with the goal of improving navigation of shared spaces on our fleet of BWIBots.
    
    \section{Conclusion}
    We have discussed work in RoboCup@Home and the Building-Wide Intelligence Project currently being carried out at UT Austin. Our work on these two projects has been synergistic. Contributions from each project have helped to push forward advances in the other, aiding us towards long-term autonomy, a high level of interactivity and utility in BWI, as well as our hopes of victory in RoboCup@Home.
    
    \section*{Acknowledgements}
    
    This work has taken place in the Learning Agents Research
    Group (LARG) at UT Austin.  LARG research is supported in part by NSF
    (IIS-1637736, IIS-1651089, IIS-1724157), Intel, Raytheon, and Lockheed
    Martin.  Peter Stone serves on the Board of Directors of Cogitai, Inc.
    The terms of this arrangement have been reviewed and approved by the
    University of Texas at Austin in accordance with its policy on
    objectivity in research.

    \bibliographystyle{aaai}

\end{document}